\documentclass[a4paper, 10pt, conference]{ieeeconf}      

\IEEEoverridecommandlockouts                              

\usepackage[utf8]{inputenc}
\usepackage{amsmath}
\usepackage{amsfonts}
\usepackage{textcomp}
\usepackage{graphicx}
\usepackage{authblk}
\usepackage{subfig}
\usepackage{gensymb}

\title{\LARGE \bf
  Space CoBot: modular design of an holonomic aerial robot for indoor microgravity environments
}

\author{Pedro Roque$^{1}$ and Rodrigo Ventura$^{1}$
\thanks{$^{1}$Pedro Roque and Rodrigo Ventura are with Institute for Systems and Robotics, Instituto Superior Técnico, Universidade de Lisboa, Av. Rovisco Pais, 1, 1049-001 Lisboa, Portugal.
        {\tt\small pedro.roque@tecnico.ulisboa.pt, rodrigo.ventura@isr.tecnico.ulisboa.pt}}%
}

\begin{document}

\maketitle
\thispagestyle{empty}
\pagestyle{empty}

\begin{abstract}

  This paper presents the design of a small aerial robot for inhabited
  microgravity environments, such as orbiting space stations (e.g.,
  ISS). In particular, we target a fleet of robots, called Space
  CoBots, for collaborative tasks with humans, such as telepresence
  and cooperative mobile manipulation. The design is modular,
  comprising an hexrotor based propulsion system, and a stack of
  modules including batteries, cameras for navigation, a screen for
  telepresence, a robotic arm, space for extension modules, and a pair
  of docking ports. These ports can be used for docking and for
  mechanically attaching two Space CoBots together. The kinematics is
  holonomic, and thus the translational and the rotational components
  can be fully decoupled. We employ a multi-criteria optimization
  approach to determine the best geometric configuration for maximum
  thrust and torque across all directions. We also tackle the problem
  of motion control: we use separate converging controllers for
  position and attitude control. Finally, we present simulation
  results using a realistic physics simulator. These experiments
  include a sensitivity evaluation to sensor noise and to unmodeled
  dynamics, namely a load transportation.

\end{abstract}

\section{INTRODUCTION}
\label{sec:introduction}

Multirotor vehicles have recently gained widespread use in various
application areas, such as transportation, surveillance, and even
entertainment. These platforms are mechanically simple and typically
lightweight, resulting in a low cost of acquisition and
maintenance. However, to the best knowledge of the authors, there is
no known application of multirotor vehicles in space
environments. Discarding the obvious impossibility of maneuvering in
the absence of air, we consider here the use of these vehicles in
human-compatible pressurised spaces under microgravity, such as inside
inhabited space stations.

The idea of aerial vehicles inside space stations is not new. The NASA
project SPHERES (Synchronized Position Hold Engage Reorient
Experimental Satellites) started in 2000 with the design of a small
pressurised air propulsion vehicle~\cite{miller00}. In 2006 three
SPHERES units were deployed aboard the International Space Station
(ISS)~\cite{nolet07}.  However, the use of pressurised air makes the
mechanical complexity significantly higher than a multirotor based
solution.  More recently, NASA proposed the Astrobee vehicle, with a
simpler propulsion system based on several centrifugal fans and
nozzles~\cite{bualat15}. The Astrobee also features a 2 DoF arm and
docking capability. However, most of its volume (about 67\%) and
area (about 56\%) are occupied by the propulsion system, where each
fan requires an unobstructed duct from side to side.

The most prominent feature of these spaces is the absence of gravity
exerted in bodies, that is, the gravity force canceled
by the orbital motion of the air mass (microgravity). We will exploit this feature in
the following way. Most multirotor orient their propellers
vertically along parallel axes. This maximizes thrust vertically in
order to compensate for the gravity force. However, in doing so they
need to tilt in order to move sideways. Moreover, most of the energy
is spent in compensating for gravity. In its absence, not
only there is no need to compensate for gravity, but also the
energy consumption is expected to be significantly less. We propose
to orient the propeller rotation axes non-parallel among them, in
such a way we obtain a fully holonomic vehicle. We expect to
increase maneuverability, being a particularly relevant feature inside
confined spaces such as the interior of a space station.

Holonomic multirotors have been proposed in the past, but the
literature is scarce. In~\cite{jiang13,voyles12} a dexterous
hexarotor has been proposed, where the holonomic kinematics is
used for dexterous manipulation. However, this vehicle is designed for
Earth applications, and thus a trade-off is necessary between dexterity
and gravity compensation. Our design is also based in an hexarotor,
but since we consider microgravity, we will design the propeller
orientation in order to maximize maneuverability along all directions.

The goal of this paper is to present a design of a aerial robot for
microgravity environments. We target a modular and multipurpose
vehicle, providing, for instance, (1)~telepresence, requiring the
vehicle to manoeuvre in such a way the screen and camera is pointed
towards the party onboard, and (2)~mobile manipulation, requiring
force closure in order to compensate the reaction force at the end
effector. Our design also includes a docking port for docking into a
charging station and stacking of multiple units for, e.g., combined
thrust.

This paper is structured as follows: the proposed design is detailed in
Section~\ref{sec:vehicle-design}, in Section~\ref{sec:propulsion} we
derive a parametric model of the propulsion system, which is then used
in Section~\ref{sec:design-parameters} for the optimization its
parameters. Section~\ref{sec:posit-attit-contr} presents a dynamical
model of the vehicle together with a
convergent controller for this holonomic vehicle, and a
preliminary validation in a realistic simulation environment is
presented in Section~\ref{sec:prel-valid}, while
Section~\ref{sec:concl-future-work} wraps up the paper with concluding
remarks and a reference for future work.

This paper builds upon and extends an early draft available in
arXiv~\cite{yoda:arxiv16}.


\section{VEHICLE DESIGN}
\label{sec:vehicle-design}

\begin{figure}
  \centering
  \subfloat[]{\includegraphics[width=0.49\linewidth]{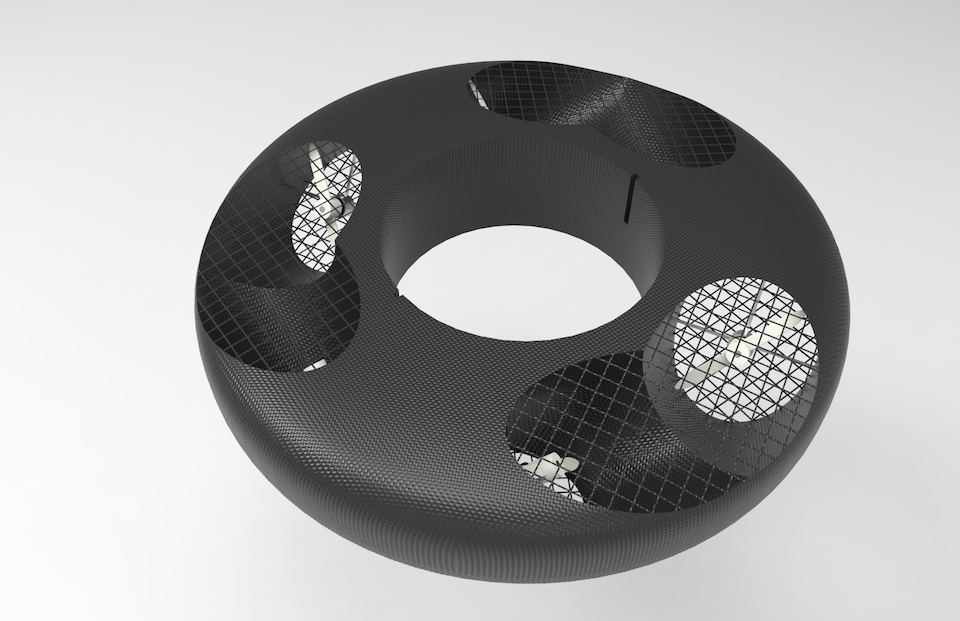}} \hfill
  \subfloat[]{\includegraphics[width=0.49\linewidth]{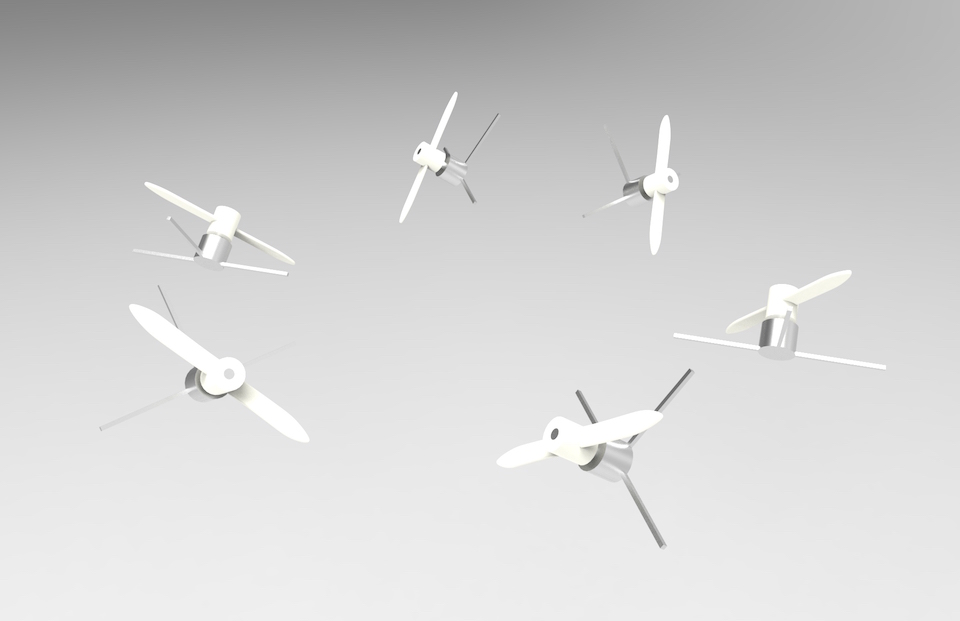}}
  \caption{Propulsion system representation: (a)~overall view of the module, and (b)~relative placement of the propellers.}
  \label{fig:propulsion_system}
\end{figure}

\begin{figure}
  \centering
  \includegraphics[width=1\linewidth]{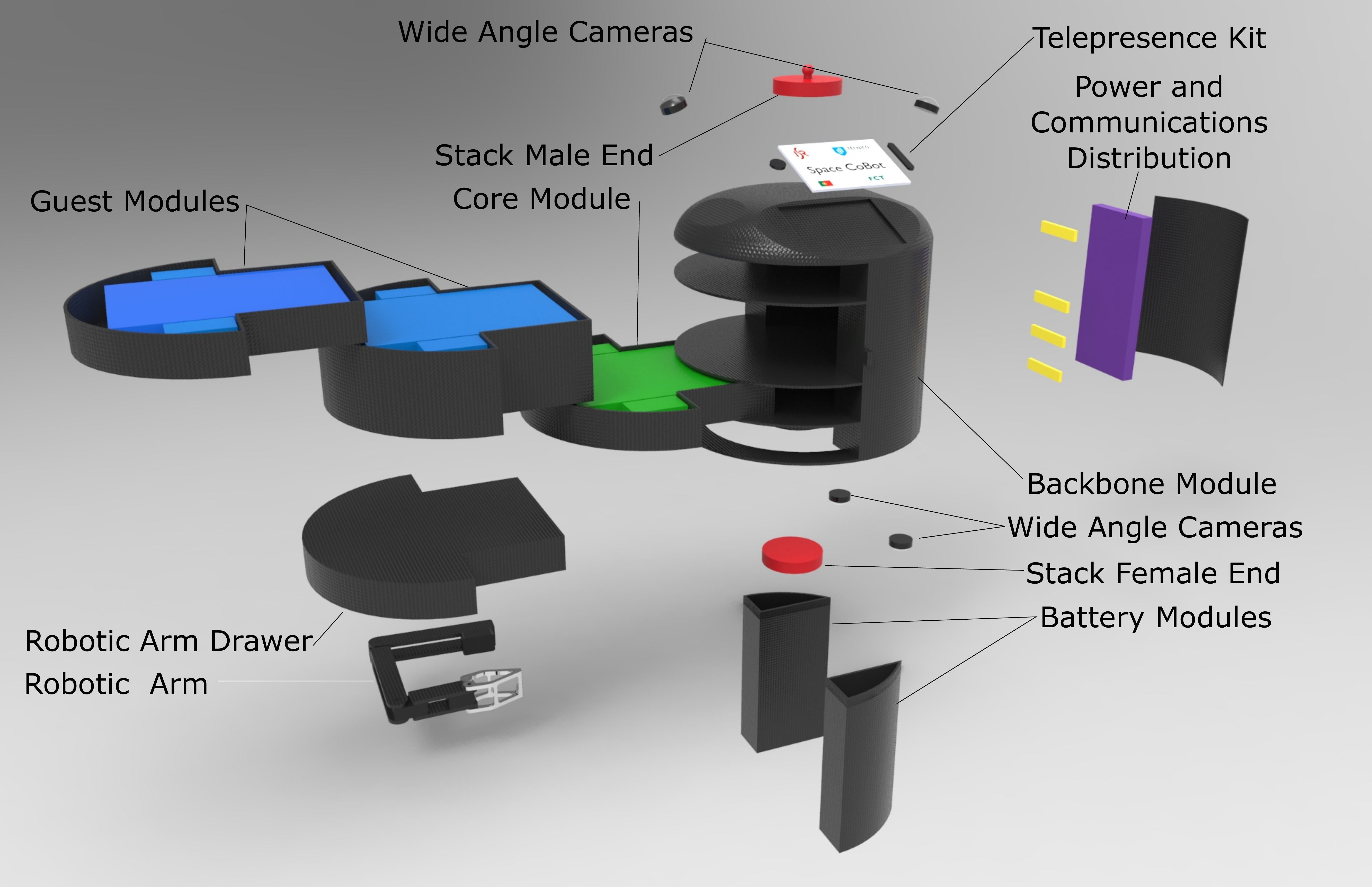}
  \caption{Overall view of Space Cobot modules.}
  \label{fig:modules}
\end{figure}

As space operation is expensive, we target a modular vehicle whose
parts can be easily maintained and that is easily extensible with
additional modules, e.g., hosting of scientific experiment
testbeds. An overview of all the modules is provided in Figure
\ref{fig:modules} and explained below.

To provide propulsion to the vehicle, we use 6 electric motors with
4'' propellers, arranged in such a way the kinematics is holonomic. We
defer a detailed description of this system, along with its
optimization, to Sections~\ref{sec:propulsion}
and~\ref{sec:design-parameters}. Our design of this system is shown in
Figure~\ref{fig:propulsion_system}.

The propulsion module attaches to the central core of the vehicle, a backbone module, that provides support for all remaining subsystems: batteries, a pair of docking connectors, cameras, extension modules, telepresence equipment, and a robotic manipulator.

For positioning inside the station we propose a vision-based
localization method employing 4 wide-angle cameras, two on the top
side of the vehicle and two on the bottom. These cameras also enable
visual servoing of the robotic arm. Autonomous navigation is
delivered by onboard computation, the Core module, capable of video
processing and propulsion actuation, as well the management of the
extension modules.

\begin{figure}
  \centering
  \subfloat[]{\includegraphics[width=0.49\linewidth]{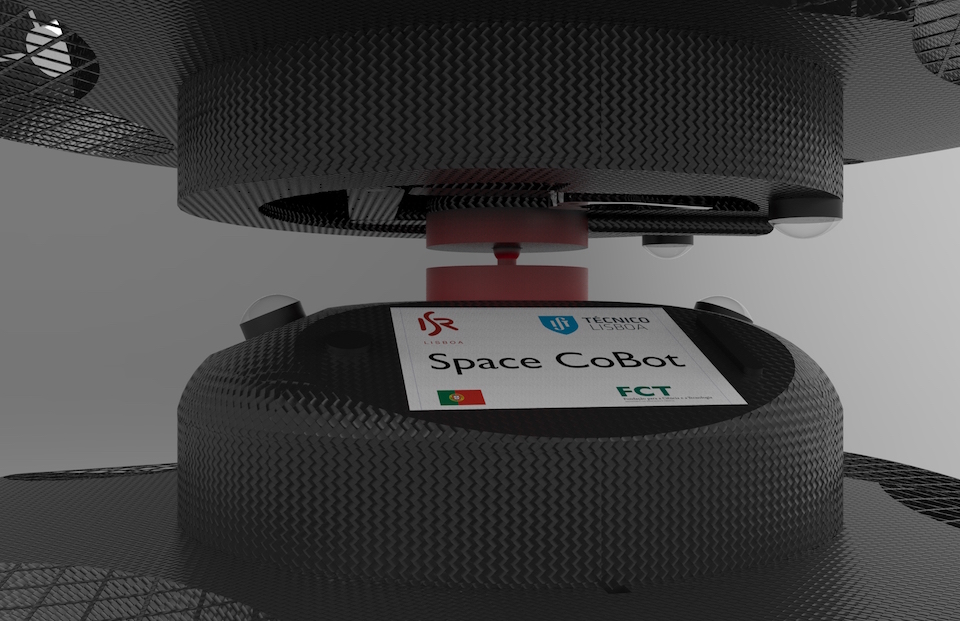}} \hfill
  \subfloat[]{\includegraphics[width=0.49\linewidth]{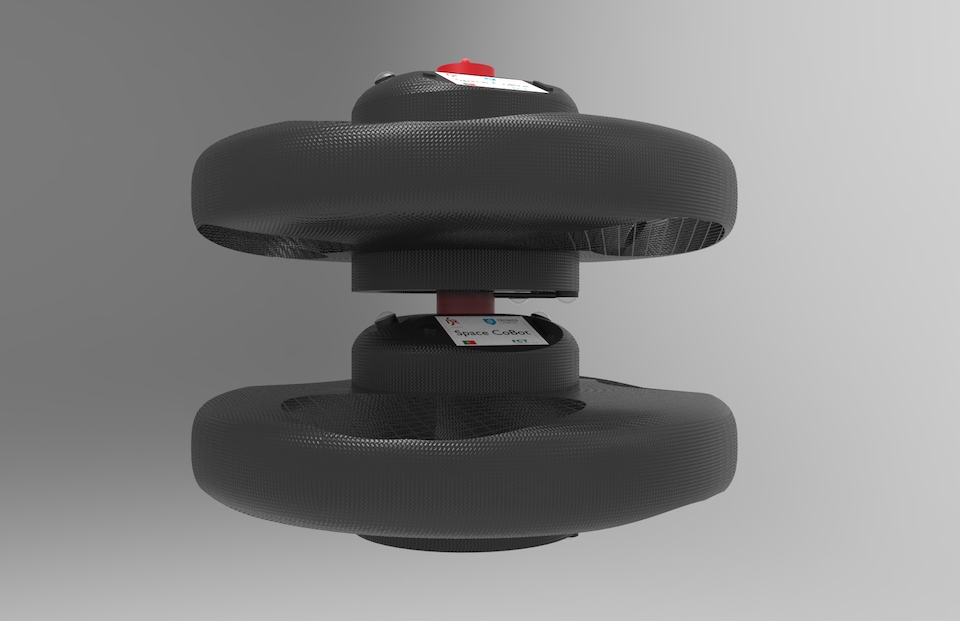}}
  \caption{Space Cobot coupling system : (a)~coupling mechanism detail , and (b)~two robots coupled together.}
  \label{fig:coupling}
\end{figure}

The docking ports provide multiple functions: they (1)~enable the
connection between the robot and a charging dock, they (2)~provide a
physical, solid connection between robots (stacking) for cooperative
tasks such as faster transportation of heavy loads. For charging, the
stacking of multiple robots also enables daisy-chained charging across several robots, thus saving space. A detailed view of the docking ports is provided on Figure~\ref{fig:coupling}.

Two possible applications of the robot are telepresence and object
manipulation. Telepresence can be performed using the onboard display,
speakers, and microphone. Our design has enough space for a common
tablet-sized display. Mobile manipulation is crucial for robot-environment interaction. A robotic arm with a gripper is provided on the bottom part of the robot, providing several manipulation capabilities such as cargo transportation. This module fits in a standard module bay and can be replaced for additional guest module space.

For safer operation, electric propulsion was used instead of pressurized gas. Along with this, we also feature electromagnetic and acoustic noise reduction methods by using grounded copper mesh on the vehicles construction and acoustic absorvig materials around the motors airflow tunnel. With this, we aim at long periods of operation aboard space stations.


\section{PROPULSION MODEL}
\label{sec:propulsion}



\begin{figure}
  \centering
  \includegraphics[width=0.6\linewidth]{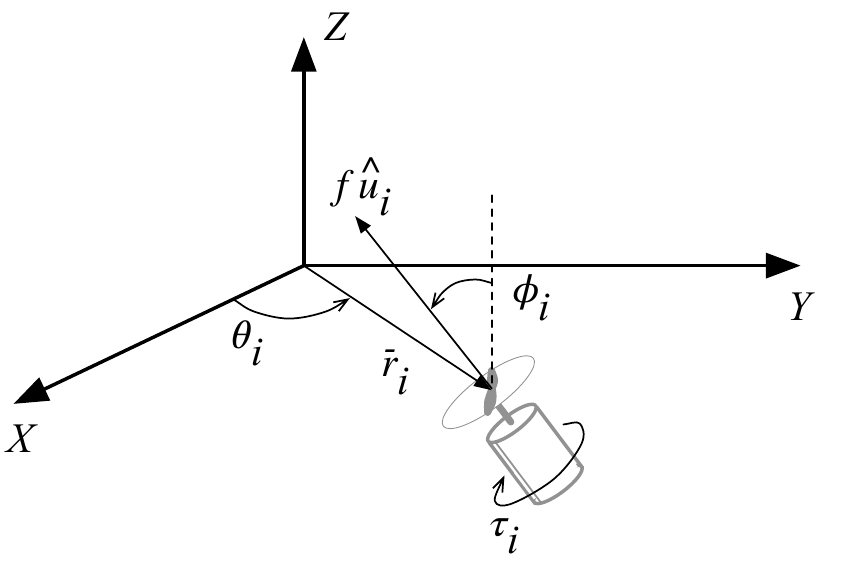}
  \caption{Notation used for modeling a single propeller with respect
    to the body frame~$\mathcal{B}$, centered on the vehicle's CoM.}
  \label{fig:prop}
\end{figure}

Consider a single propeller $i$ whose motor is rigidly linked to the
body frame~$\mathcal{B}$, as depicted in Figure~\ref{fig:prop}.  This
propeller  generates a reaction force (thrust) $\bar{F}_i$ and  torque
$\bar{M}_i$ on the vehicle body.
While the former results directly from the
propeller thrust,
\begin{align}
  \label{eq:Fi}
  \bar{F}_i &= f_i \hat{u}_i \\
  f_i &= K_1 u_i
\end{align}
where $u_i$ is the actuation signal and $f_i$ is the scalar thrust,
the later is the sum of two components: the torque caused by the
non-central thrust
and the propeller reaction torque:
\begin{align}
  \label{eq:Mi}
  \bar{M}_i &= \bar{r}_i\times\bar{F}_i - \tau_i \hat{u}_i \\
  \tau_i &= w_i K_2 u_i
\end{align}
where $\times$ denotes vector cross product and $w_i$ is $-1$ or $1$
depending on whether the propeller rotates clockwise or anti-clockwise
for a positive forward thrust $f_i>0$.  These linear relationships
result from the momentum-blade element theory by considering $u=n^2$,
where $n$ is the blade's rotation speed (in revolutions per second) while the
constants $K_1$ and $K_2$ given by
\begin{equation}
  \label{eq:blade}
  K_1 = \rho D^4 C_T \qquad K_2 = \frac{\rho D^5}{2\pi} C_P
\end{equation}
where $\rho$ is the air density, $D$ is the propeller diameter, and
the $C_T$ and $C_P$ are blade dependent adimensional constants called
thrust and power coefficients~\cite{mccormick95}.

The relative position $\bar{r}_i$ of the $i$-th propeller, orthogonal to the
$Z$ axis, and the unit vector $\hat{u}_i$, aligned with the propeller
axis, are uniquely defined by the angles $\theta_i$ and $\phi_i$, and can be
easily obtained from geometrical reasoning by:
\begin{equation}
  \bar{r}_i = \left(
    \begin{array}{c}
      d\cos(\theta_i) \\
      d\sin(\theta_i) \\
      0
      \end{array}
    \right)
    \qquad
    \hat{u}_i = \left(
      \begin{array}{c}
        \sin(\theta_i)\sin(\phi_i) \\
        -\cos(\theta_i)\sin(\phi_i) \\
        \cos(\phi_i)
      \end{array}
      \right)
\end{equation}
where $d=\|\bar{r}_i\|$ is the distance from the propeller to the CoM.
Stacking together the resulting force $\bar{F}_i$ and torque
$\bar{M}_i$, one can obtain the linear relation
\begin{equation}
  \label{eq:fm}
  \left(
    \begin{array}{c}
      \bar{F}_i \\
      \bar{M}_i
    \end{array}
  \right) = \bar{a}_i\,u_i
\end{equation}
where
\begin{equation}
  \begin{split}
    \bar{a}_i  &= \left(
      \begin{array}{c}
        K_1 \hat{u}_i \\
        K_1 \bar{r}_i \times \hat{u}_i - w_i K_2 \hat{u}_i
      \end{array}
    \right) \\
    &= \left(
      \begin{array}{c}
        K_1 \sin(\theta_i)\sin(\phi_i) \\
        -K_1 \cos(\theta_i)\sin(\phi_i) \\
        K_1 \cos(\phi_i) \\
        \left[K_1 d \cos(\phi_i) - w_i K_2 \sin(\phi_i)\right]\sin(\theta_i) \\
        -\left[K_1 d \cos(\phi_i) - w_i K_2 \sin(\phi_i)\right]\cos(\theta_i) \\
        -K_1 d \sin(\phi_i) - w_i K_2 \cos(\phi_i)
      \end{array}
    \right)
    \end{split}
\end{equation}

For a vehicle with $N$ propellers, the resulting force and torque will
be given by the sum of the contributions of each propeller, given with
respect to the CoM, by~\eqref{eq:fm}. This sum can be put in matrix
form as
\begin{equation}
  \label{eq:au}
  \left(
    \begin{array}{c}
      \bar{F} \\
      \bar{M}
    \end{array}
  \right) = \mathbf{A}\,\bar{u}
\end{equation}
where $\mathbf{A}=[\bar{a}_1 \cdots \bar{a}_N]$ is a square matrix,
hereby called \emph{actuation matrix}, and $\bar{u}=[u_1\cdots u_N]^T$
is the actuation input vector. \emph{The crucial observation is that, if the
actuation matrix $\mathbf{A}$ has at least rank 6, the linear
equation~\eqref{eq:au} can be solved for $\bar{u}$ for any given
combination of $\bar{F}$ and $\bar{M}$.}  A necessary (but not
sufficient\footnote{Sufficiency required $\mathbf{A}$ to be full
  rank.}) condition for this to be true is to have at least 6
propellers, thus justifying the hexarotor design considered in this
paper. Note that this matrix only depends on the vehicle's design
parameters, defined by the angles $\{\theta_i\}$ and $\{\phi_i\}$,
distance $d$, the trust coefficients $K_1$, $K_2$, and $\{w_i\}$.


\section{OPTIMIZATION OF THE DESIGN PARAMETERS}
\label{sec:design-parameters}

In this section we will study how the design parameters translate to
the actuation matrix $\mathbf{A}$ defined in the previous section. In
particular, we will consider how actuation limits translates into
maximum values of force and torque. We will then optimize these
parameters in order to maximize the upper bound of forces and torques,
over all directions, imposed by a bounded actuation.

\subsection{Problem statement}
\label{sec:problem-statement}


We will start by considering that each actuation signal is
bounded between $-1$ and $1$, that is,
\begin{equation}
  \label{eq:ahcube}
  -1 \leq u_i \leq 1 \quad \text{for} \quad i=1,\ldots,6
\end{equation}
According to \eqref{eq:au}, this hypercube will map onto a
6-dimensional convex polyhedron\footnote{A convex polyhedron is an
  intersection of a finite number of half-spaces~\cite{jeter86}.} in
the $(\bar{F}, \bar{M})$ space.  Any other choice of bounds is
possible by appropriately scaling constants $K_1$ and $K_2$.  However,
it assumes that the maximum propeller trust is symmetric with respect
to the direction of rotation. The remaining parameters are the angles
$\{\phi_i\}$. We will base our analysis on the optimization of these
angles with respect to various criteria.

Our goal will be to find the configurations of angles $\{\phi_i\}$
that maximize the range of forces (and torques) over all
directions.
Geometrically, this corresponds to changing $\{\phi_i\}$
such that a ball of nonzero radius can fit inside the 3-dimensional
convex polyhedron in the $\bar{F}$ space mapped by the actuation
hypercube in~\eqref{eq:ahcube}, while keeping zero torque,
$\bar{M}=0$. A similar reasoning applies to the torque space
$\bar{M}$, while keeping $\bar{F}=0$. 

First, we will address the problem of computing the maximum force
along a given direction specified as a unit vector $\hat{e}$, while
maintaining a zero torque.  From~\eqref{eq:au}, and assuming that
$\mathbf{A}$ is full rank, we get
\begin{equation}
  \bar{u} = \mathbf{A}^{-1}
  \left(
    \begin{array}{c}
      \bar{F} \\
      \bar{M}
    \end{array}
  \right) =\mathbf{A}^{-1}
  \left(
    \begin{array}{c}
      F \hat{e} \\
      0
    \end{array}
  \right)
\end{equation}
where $F>0$ is the force magnitude. For what follows, it will be
convenient to express the $\mathbf{A}^{-1}$ matrix as blocks of three
dimensional row vectors:
\begin{equation}
  \label{eq:bici}
  \mathbf{A}^{-1} = \left[
    \begin{array}{cc}
      b_1^T & c_1^T \\
      \vdots & \vdots \\
      b_6^T & c_6^T \\
    \end{array}
    \right]
\end{equation}
where $b_i,c_i\in\mathbb{R}^3$. Then, the actuation of the $i$-th
propeller is given by
\begin{equation}
  u_i = F b_i^T \hat{e}
\end{equation}
Since $|u_i|\leq1$, we have $F|b_i^T\hat{e}|\leq1$, and thus $F$ has
this upper bound:
\begin{equation}
  F \leq \frac{1}{|b_i^T\hat{e}|}
\end{equation}
Since this inequality has to be satisfied for all propellers
$i=1,\ldots,6$, the maximum force $F^\mathrm{max}_{\hat{e}}$ is given
by the lowest of these upper bounds
\begin{equation}
  F^\mathrm{max}_{\hat{e}} = \min_i \frac{1}{|b_i^T\hat{e}|}
\end{equation}
This force is the maximum force along a given direction $\hat{e}$. The
maximum force attainable in any direction can be obtained by
minimising this force over all possible directions. Since
$|b_i^T\hat{e}|\leq\|b_i\|$, this minimum is given by
\begin{equation}
  \label{eq:fmax}
  F^\mathrm{max} = \min_i \frac{1}{\|b_i\|}
\end{equation}
The same reasoning can be applied to the torques: consider a torque
$\bar{M}=M\hat{e}$ along an arbitrary direction defined by $\hat{e}$,
the corresponding actuation with $\bar{F}=0$ is $u_i=M c_i^T \hat{e}$,
resulting in the following maximum torque along any direction:
\begin{equation}
  \label{eq:mmax}
  M^\mathrm{max} = \min_i \frac{1}{\|c_i\|}
\end{equation}

Now, these maximum forces and torque values depend on the design
parameters. In the following we will use an optimization approach to
find the values of these parameters that maximize the maximum force
and/or torque. We will consider the propellers to be equally
distributed radially, that is,
\begin{equation}
  \label{eq:thetas}
  \theta_i = (i-1) \frac{\pi}{3}
\end{equation}
and a fixed distance $d$, as well as the constants $K_1$ and $K_2$. All the
remaining parameters will be the unknown variables:
\begin{equation}
  \bar{\psi} = ( \phi_1, \ldots, \phi_6, w_1,\ldots, w_6 )^T
\end{equation}
with the feasibility domain defined by
\begin{equation}
  \Psi = \left\{ \bar{\psi} \::\: |\phi_{1,\ldots,6}| \leq
  \phi_{max}, w_{1,\ldots,6}\in\{-1,1\} \right\}
\end{equation}
where $\phi_{max}$ is the maximum allowed deviation from the
vertical. As a mechanical constraint to allow obstructionless air flow
we considered $\phi_{max}=\pi/3$ in this work.

We can restate the maximization of~\eqref{eq:fmax}
and~\eqref{eq:mmax} using the epigraph form~\cite{boyd04}, thus getting rid of the
maximization of a minimum:
\begin{equation}
  \label{eq:minp}
  \begin{split}
    &\text{minimize}\: p \\
    &\text{subject to:} \\
    &\quad p \geq \|b_i\|^2, i=1,\ldots,6
  \end{split}
\end{equation}
for the force and
\begin{equation}
  \label{eq:minq}
  \begin{split}
    &\text{minimize}\: q \\
    &\text{subject to:} \\
    &\quad q \geq \|c_i\|^2, i=1,\ldots,6
  \end{split}
\end{equation}
for the torque, where $\{b_i\}$ and $\{c_i\}$ depend non-linearly on
the parameters $\bar{\psi}$ through the inverse of the actuation
matrix $\mathbf{A}$ as~\eqref{eq:bici}. In this form, the optimization
variables are augmented with the cost, that is,
$\bar{\psi}_p=(p, \bar{\psi})$ for the problem~\eqref{eq:minp} and
$\bar{\psi}_q=(q, \bar{\psi})$ for~\eqref{eq:minq}.  It can be readily
seen that these forms maximize~\eqref{eq:fmax} and~\eqref{eq:mmax},
where the resulting maximum forces and torques can be recovered using
$F^{max}=1/\sqrt{p}$ and $M^{max}=1/\sqrt{q}$.

\subsection{Multi-criteria optimization}
\label{sec:multi-crit-optim}

Since we intend to both maximize force and torque, we will make the
trade-off between the two explicit by taking a multi-criteria
optimization approach:
\begin{equation}
  \label{eq:multi}
  \begin{split}
    &\text{minimize}\:(p, q) \\
    &\text{subject to:} \\
    &\quad p \geq \|b_i\|^2, i=1,\ldots,6 \\
    &\quad q \geq \|c_i\|^2, i=1,\ldots,6
  \end{split}
\end{equation}
In this problem, the optimization variables are augmented with both
$p$ and $q$, $\bar{\psi}_{pq}=(p, q, \bar{\psi})$, and we have two
cost functions, say $J_1(\bar{\psi}_{pq})=p$ and
$J_2(\bar{\psi}_{pq})=q$. The solution of this multi-criteria
optimization problem is the set $\mathcal{P}$ of non-dominated
solutions, defined by: $\bar{\psi}_{pq}^0\in P$ if and only if there is no
$\bar{\psi}_{pq}\in\Psi$ such that
$J_i(\bar{\psi})\leq J_i(\bar{\psi}^0)$ for all $i\in\{1,2\}$ and
$J_i(\bar{\psi})< J_i(\bar{\psi}^0)$ for at least one
$i\in\{1,2\}$. This set is also called \emph{Pareto optimal
  set}~\cite{statnikov95}, a subset of the \emph{objective space}
defined by all~$(J_1(\bar{\psi}),J_2(\bar{\psi}))$ for
$\bar{\psi}\in\Psi$.

Apart from very simple cases, the Pareto optimal set is not trivial to
obtain exactly. Thus, we will make a pointwise approximation using the
Normally Boundary Intersection (NBI) method~\cite{das98}. This method
is guaranteed to obtain Pareto optimal points if the objective space
is convex. But it is still capable of obtaining points in ``sufficiently
concave'' parts of the objective space~\cite{das98}.

The first step of NBI is to obtain the minimizers of the each cost
function taken individually. These are also called \emph{shadow
  minima}. Let us start by considering the first minimization
problem~\eqref{eq:minp}, where $\bar{\psi}^*_p$ is the minimizer with
minimum cost $p^*$.  Then, this minimizer both minimizes $p$
in~\eqref{eq:multi} and, together with $q^0=\max_i\,\|c_i\|^2$, is a
non-dominated solution of~\eqref{eq:multi}, and thus belongs to its
Pareto optimal set. This results from the fact that this $q^0$ is the
smallest one that still satisfies the constrains of~\eqref{eq:multi}:
any $(p^*,q)$ with $q>q^0$ is dominated by $(p^*,q^0)$.  The same
reasoning can be applied to~\eqref{eq:minq}, resulting in the
minimizer $\bar{\psi}^*_q$, with minimum cost $q^*$, that together
with $p^0=\max_i\,\|b_i\|^2$ is also a non-dominated solution
of~\eqref{eq:multi}, thus also belonging to its Pareto optimal set.

On the $(p,q)$ space, these two non-dominated solutions corresponds
to two extremal points, $(p^*,q^0)$ and $(p^0,q^*)$, of the Pareto
optimal set: no feasible solutions exists neither to the left of
$p^*$ nor lower than $q^*$. \emph{The application of NBI to a two cost
function problem amounts to scanning along a straight line joining
$(p^*,q^0)$ and $(p^0,q^*)$, and then, for each point on this
straight line, to determine the single non-dominated solution along
the orthogonal direction.}

\begin{figure}
  \centering
  \includegraphics[width=0.6\linewidth]{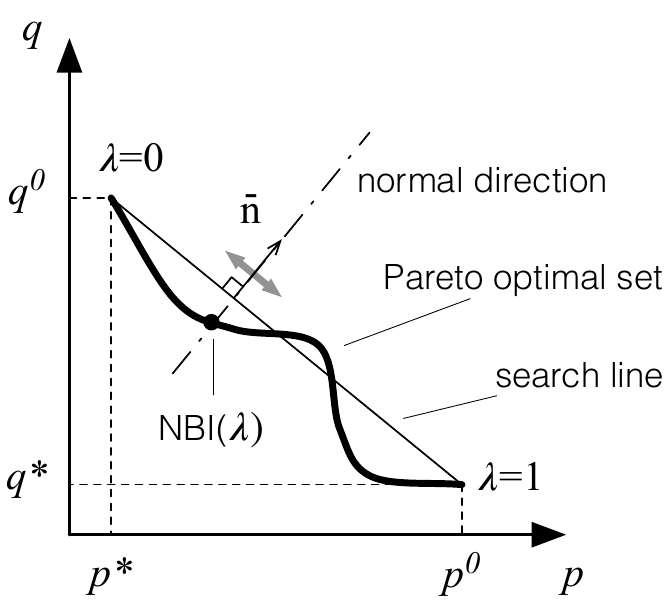}
  \caption{Illustration of the NBI method: given a point defined by
    $\lambda\in[0;1]$, along the search line between $(p^*,q^0)$ and
    $(p^0,q^*)$, the optimization is done along the normal direction
    spawned by vector $\bar{n}$, resulting on the $NBI(\lambda)$
    intersection point.}
  \label{fig:nbi}
\end{figure}

Figure~\ref{fig:nbi} illustrates the NBI method. This straight line
can be parametrized by a $\lambda$ value ranging between 0 and 1,
resulting in $(1-\lambda)(p^*,q^0)+\lambda\,(p^0,q^*)$. An
orthogonal direction to this straight line is spawned by the vector
$\bar{n}=(q^0-q^*,p^0-p^*)$. Using again the epigraph form, but now
along this vector, we obtain the following constrained optimization
problem:
\begin{equation}
  \label{eq:nbi}
  \begin{aligned}
    &\text{minimize}\: t \\
    &\text{subject to:} \\
    &\quad (q^0-q^*)\,t + (1-\lambda) p^* + \lambda p^0 &\geq
    \|b_i\|^2 \\
    &\quad (p^0-p^*)\,t + \lambda q^* + (1-\lambda) q^0 &\geq
    \|c_i\|^2 \\
    &\qquad \text{for } i=1,\ldots,6
  \end{aligned}
\end{equation}
with the augmented vector $\bar{\psi}_t=(t, \bar{\psi})$ as
optimization variable. For a given $\lambda\in[0;1]$, the solution of
this optimization problem yields a minimizer $\bar{\psi}^*_t(\lambda)$
from which the corresponding point in the $(p,q)$ space is
\begin{equation}
  NBI(\lambda) = (p_\lambda, p_\lambda), \quad p_\lambda = \min_i \|b_i\|^2, \quad q_\lambda = \min_i \|c_i\|^2
\end{equation}
from which the maximum values of force and torque can be recovered as
above mentioned.

Since these problems cannot be solved in closed form, we will make use
of numerical optimization methods. The following section presents the
numerical results obtained for this problem.

\subsection{Numerical results}
\label{sec:numerical-results}

The optimization problem in~\eqref{eq:nbi} shows some features that
make it non-trivial to solve: it is both strongly non-convex with
mixed continuous and discrete variables. First, we will factor out the
discrete part by iteratively trying each combination of $\{w_i\}$
values modulo rotations (also called \emph{orbits}\footnote{For
  instance, $[1,-1,1,1,1,1]$ and $[1,1,-1,1,1,1]$ belong to the same
  orbit, and thus it is redundant to try both of them.}): from its
$2^6=64$ possible combinations, only 14 correspond to combinations
where no pair can be made equal after rotating one of them. Second, we
use a random multistart initialization together with a convex
optimization algorithm: for each sample drawn uniformly from the
$\{\phi_i\::\:|\phi_i| \leq \phi_{max}\}$ cube, we run the Constrained
Optimization BY Linear Approximation (COBYLA) algorithm~\cite{powell94},
as implemented in the SciPy optimization package.


To make the relation between force and actuation dimensionless, we
divided the actuation matrix by $K_1$. This way, the only dependence
on physical coefficients of this matrix is on the parameters $d$ and
the ratio $K_2/K_1$. Using~\eqref{eq:blade}, this ratio can be
expressed as
\begin{equation}
  \frac{K_2}{K_1} = \frac{D}{2\pi} \frac{C_P}{C_T}
\end{equation}
For various small propellers (of about 4'') we found\footnote{We used the
  UIUC Propeller Data Site, Vol. 2,
  http://m-selig.ae.illinois.edu/props/propDB.html (retrieved
  Nov-2015).} this ratio to be approximately~0.01. And thus we used
this value to obtain the numerical results presented in this section.
For $d$ we used the one from the design presented in
Section~\ref{sec:vehicle-design}, that is, $d=0.16$.


The extremes of the NBI search line are the shadow minima,
\textit{i.e.,} the minima of~\eqref{eq:minp}
and~\eqref{eq:minq}. For~1000 random initializations, we obtained
these values for the shadow minima: $(p^*,q^0)=(0.250, 15.01)$ and
$(p^0,q^*)=(0.4167,9.728)$.


\begin{figure*}
  \centering
  \subfloat[]{\includegraphics[height=0.3\linewidth]{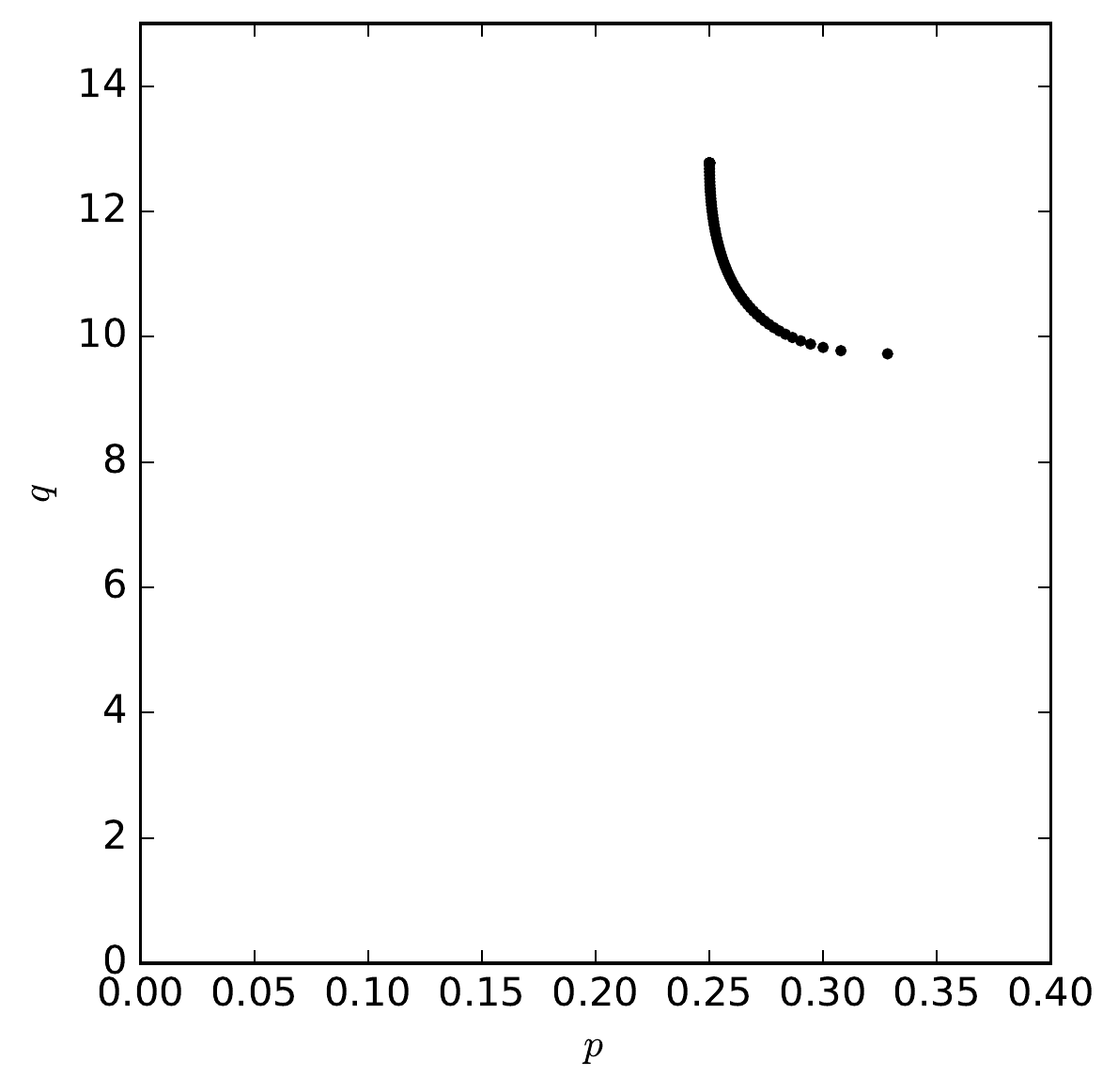}} \hfill
  \subfloat[]{\includegraphics[height=0.3\linewidth]{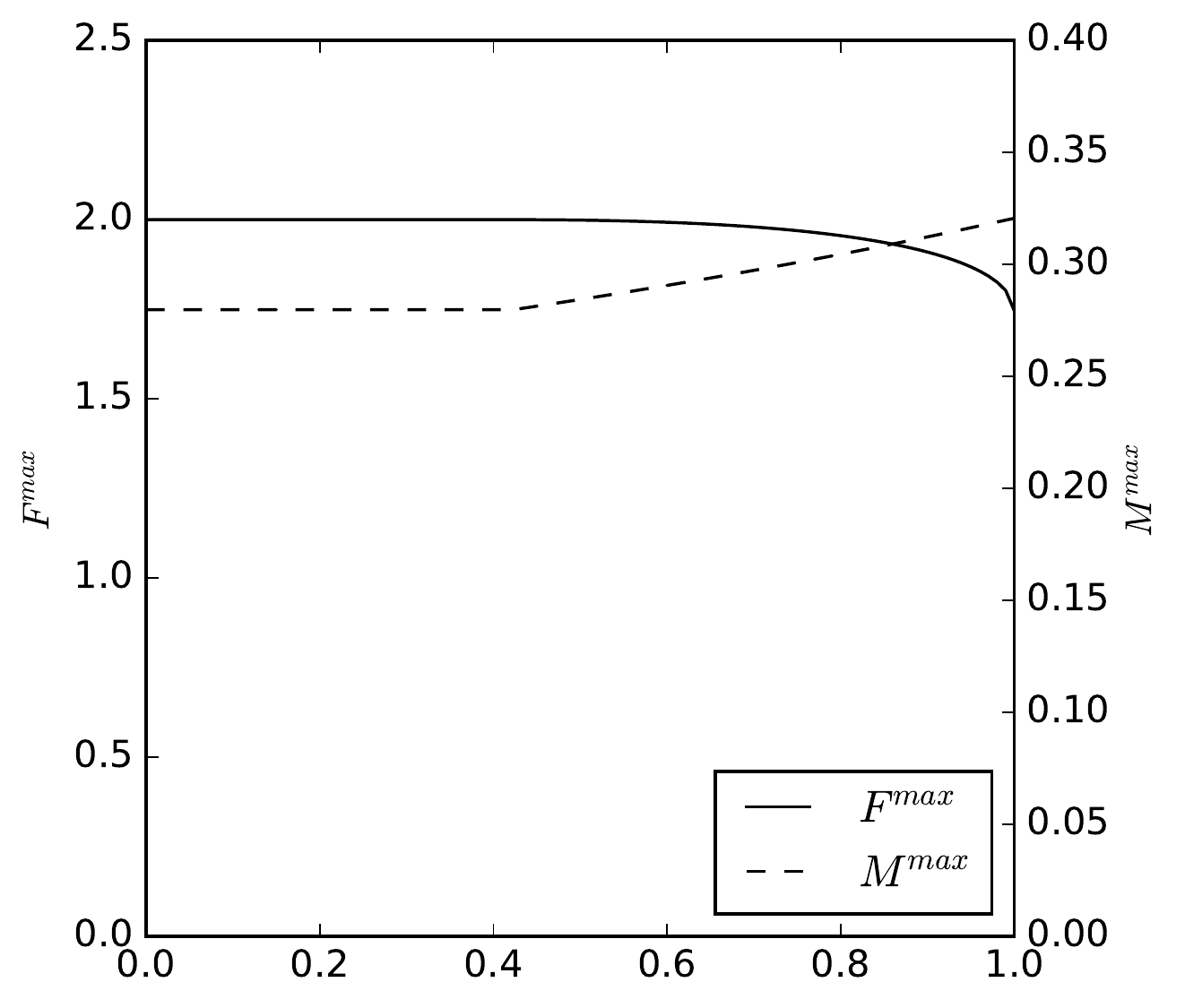}} \hfill
  \subfloat[]{\includegraphics[height=0.3\linewidth]{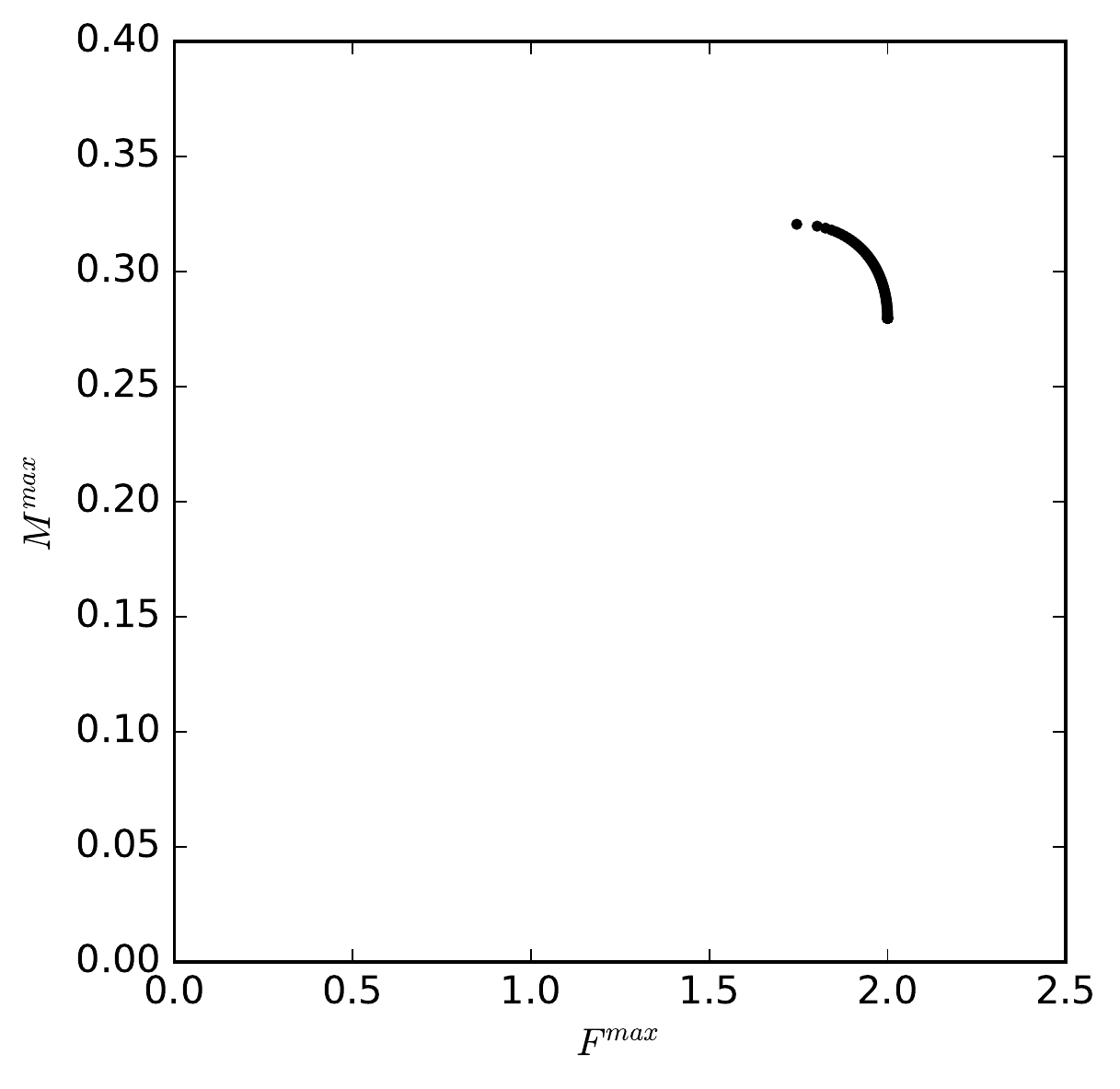}}
  \caption{Pointwise approximation to the Pareto optimal set using the
    NBI method: (a)~obtained points in the $(p.q)$ space,
    (b)~$F^{max}$ and $M^{max}$ in function of $\lambda$, and (c)~in
    the $(F^{max},M^{max})$ space. The dimensions for $F^{max}$ and
    $M^{max}$ have no physical meaning because of the division of the
    actuation matrix by~$K_1$, as explained in the text, and thus are
    to be understood in relative terms only.}
  \label{fig:nbiout2}
\end{figure*}

\begin{table*}
  \centering
  \begin{tabular}{l|cccccc|rrrrrr|cc}
    $\lambda$
    & $\phi_1$ & $\phi_2$ & $\phi_3$ & $\phi_4$ & $\phi_5$ & $\phi_6$
    & $w_1$ & $w_2$ & $w_3$ & $w_4$ & $w_5$ & $w_6$
    & $F^{max}$ & $M^{max}$ \\
    \hline
    0
    & 54.74 &-54.73 & 54.74 &-54.74 & 54.73 &-54.74
    &-1 & 1 &-1 & 1 &-1 & 1
    & 2.000 & 0.2798 \\
    0.25
    & 54.73 &-54.74 & 54.74 &-54.73  & 54.74 &-54.74
    &-1 & 1 &-1 & 1 &-1 & 1
    & 2.000 & 0.2798 \\
    0.5
    & 53.73 &-53.73 & 53.73 &-53.73 & 53.73 &-53.73
    &-1 & 1 &-1 & 1 &-1 & 1
    & 1.999 & 0.2844 \\
    0.75
    & 49.56 &-49.56 & 49.56 &-49.56 & 49.56 &-49.56
    &-1 & 1 &-1 & 1 &-1 & 1
    & 1.969 & 0.3009 \\
    1
    & 38.84 &-38.85 & 38.84 &-38.84  & 38.85 &-38.84
    &-1 & 1 &-1 & 1 &-1 & 1
    & 1.745 & 0.3206 \\
    \hline
  \end{tabular}
  \caption{Some of the configurations obtained for 5 equally spaced
    values of $\lambda$. The values for $\{\phi_i\}$ are shown in
    degrees. As before, the dimensions for $F^{max}$ and
    $M^{max}$ have no physical meaning.}
  \label{tab:configs}
\end{table*}

With these values, we ran our optimization method for $\lambda$
ranging from~0 to~1 on 0.01 steps, for 1000 random initializations
each. The result is a set of points approximating the Pareto optimal
set, shown in Figure~\ref{fig:nbiout2}. The (a) plot of this figure
suggests a convex Pareto front. For a range of $\lambda$ values from 0
to about 0.4, the $(p,q)$ values are constant. As $\lambda$ increases
over 0.4, there is a drop on the values of $q$, meaning a slight
increase on $M^{max}$


\begin{table}
  \centering
  \begin{tabular}{ccccccc}
    propeller ($i$) & 1 & 2 & 3 & 4 & 5 & 6 \\
    \hline
    $\theta_i$ & 0 & 60 & 120 & 180 & 240 & 300 \\
    $\phi_i$ & 55 & -55 & 55 & -55 & 55 & -55 \\
    $w_i$ & -1 & 1 & -1 & 1 & -1 & 1 \\
    \hline
  \end{tabular}
  \caption{Design parameters of the selected solution. Both
    $\{\theta_i\}$ and $\{\phi_i\}$ are expressed in degrees.}
  \label{tab:selected}
\end{table}

Table~\ref{tab:configs} shows some of the optimal configurations
obtained for some values of $\lambda$. In our choice we decided to
prefer maximum force, and thus we selected a configuration found in
the lower range of $\lambda$ values. We rounded off the angle values
to the closest integer degree, resulting in the configuration shown in
Table~\ref{tab:selected}. All of the following results shown in this
paper employ this selected configuration.

\section{POSITION AND ATTITUDE CONTROL}
\label{sec:posit-attit-contr}

The dynamical model of the vehicle, in the absence of gravity, can be
derived from the Newton and Euler equations of motion. Let us denote the position and velocity of the
body frame $\mathcal{B}$ with respect to the inertial frame
$\mathcal{I}$ as $\bar{x}$ and $\bar{v}$, the rotation matrix of frame
$\mathcal{B}$ with respect to $\mathcal{I}$ as $\mathbf{R}$, and the
angular velocity of the vehicle in the body frame $\mathcal{B}$ as
$\bar{\omega}$. Then,
\begin{equation}
  \label{eq:dynsys}
  \left\{
    \begin{aligned}
      \dot{\bar{x}} &= \bar{v} \\
      m \dot{\bar{v}} &= \mathbf{R} \bar{F} \\
      \dot{\mathbf{R}} &= \mathbf{R}\,\mathbf{S}(\bar{\omega}) \\
      \mathbf{J} \dot{\bar{\omega}} &= \bar{M} - \bar{\omega}\times\mathbf{J} \bar{\omega}
    \end{aligned}
  \right.
\end{equation}
where the constants $m$ and $\mathbf{J}$
are the vehicle's mass and moment of inertia, while
$\mathbf{S}(\bar{\omega})$ is the skew-symmetric matrix defined by
\begin{equation}
  \label{eq:S}
  \mathbf{S}(\bar{\omega}) =
  \left[
    \begin{array}{ccc}
      0 & -\omega_z & \omega_y \\
      \omega_z & 0 & -\omega_x \\
      -\omega_y & \omega_x & 0 \\
    \end{array}
  \right]
\end{equation}
for $\bar{\omega}=[\omega_x\:\omega_y\:\omega_z]^T$.

The approach used for the motion control of the vehicle exploits its
holonomic design by
decoupling the translational and rotational modes. To do so, we first
apply feedback linearisation~\cite{sastry99} to the translational part
of~\eqref{eq:dynsys}:
\begin{equation}
  \label{eq:fctrl}
    \left\{
    \begin{split}
      \dot{\bar{x}} &= \bar{v} \\
      \dot{\bar{v}} &= \bar{p} \\
      \bar{F} &= m \mathbf{R}^T \bar{p}
    \end{split}
  \right.
\end{equation}
and then design a feedback controller for $\bar{p}$. Since this
dynamical system is diagonal and second order, a PD controller is
enough to ensure exponential
convergence:
\begin{equation}
  \label{eq:pctrl}
  \begin{split}
    \bar{e}_x &= \bar{x} - \bar{x}_d \\
    \bar{e}_v &= \bar{v} - \bar{v}_d \\
    \bar{p} &= -k_x \bar{e}_x -k_v \bar{e}_v \\
  \end{split}
\end{equation}
where $\bar{x}_d$ and $\bar{v}_d$ are the desired position and
velocity vectors in the inertial frame~$\mathcal{I}$, and $k_x$ and
$k_v$ are the proportional and derivative gains of the PD controller.

For the attitude control we follow the exponentially convergent $SO(3)$ controller
proposed in~\cite{lee12}:
\begin{equation}
  \label{eq:mctrl}
  \begin{aligned}
    \bar{e}_R &= \frac{1}{2 \sqrt{1+tr[\mathbf{R}^T_d\mathbf{R}]} } S^{-1}( \mathbf{R}^T_d \mathbf{R} -
    \mathbf{R}^T \mathbf{R}_d ) \\
    \bar{e}_\omega &= \bar{\omega} - \mathbf{R}^T
    \mathbf{R}_d\,\bar{\omega}_d \\
    \bar{M} &= -k_R\,\bar{e}_R - k_\omega\,\bar{e}_\omega \\
    &+S^{-1}(\mathbf{R}^T\mathbf{R}_d\bar{\omega}_d)\mathbf{J}\mathbf{R}^T\mathbf{R}_d\bar{\omega}_d 
    +\mathbf{J}\mathbf{R}^T\mathbf{R}_d\dot{\bar{\omega}}_d \\
  \end{aligned}
\end{equation}
where $\mathbf{R}_d$ is the rotation matrix of the desired attitude,
$\bar{\omega}_d$ is the desired angular velocity vector, with $k_R$ and
$k_\omega$ as controller gains. The $S^{-1}$ function performs the inverse
operation as the one defined in~\eqref{eq:S}, that is, recovers
the vector from a given skew-symmetric matrix.


\section{SIMULATION RESULTS}
\label{sec:prel-valid}

In order to validate of the proposed design, we simulated a model of our robot with V-REP simulation framework~\cite{vrep13}. Here, we kept the size of the vehicle and estimated the real mass and inertia of the fully loaded robot. The design parameters used are the ones shown in Table~\ref{tab:selected}.  The vehicle mass is 6.05kg. In V-REP we simulated the force and torque produced by each propeller as modeled in~\eqref{eq:Fi} and~\eqref{eq:Mi}. Thus, this simulation includes the validation of both the design discussed in Section~\ref{sec:design-parameters} and the controller proposed in Section~\ref{sec:posit-attit-contr}.

The validation was divided in two phases: on the first, each motion mode (translation and attitude) is evaluated separately, and on the second, we simulated a path defined by a sequence of waypoints with attitude setpoints and a unmodeled payload transportation, thus combining both modes simultaneously. In the last case we added noise to the pose with a variance of 2cm on position, 2cm/s on velocity, 5\degree \ on attitude and 1\degree/s on angular velocity. 

\subsection{Individual motion mode}

\begin{figure}
  \centering
  \includegraphics[width=0.8\linewidth]{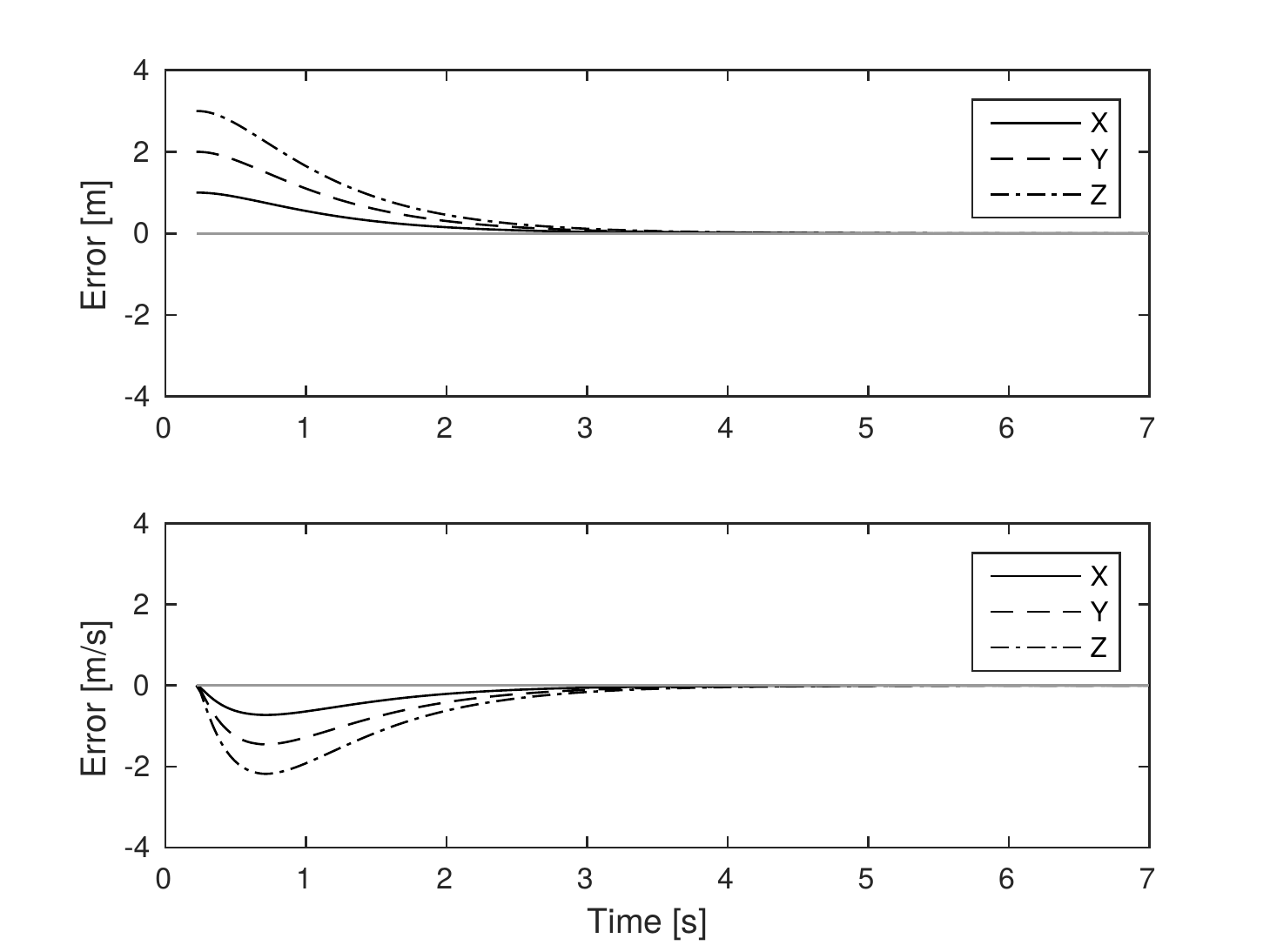}
  \caption{Translational mode error convergence, from top to bottom: $e_x$ and $e_v$.}
  \label{fig:position_errors}
\end{figure}

For this validation, the robot is holding on position $(0,0,1)$, while the reference is set at $(1,2,4)$. The total movement of the robot is 1, 2 and 3 meters along X, Y and Z axis. On Figure~\ref{fig:position_errors} we provide the obtained results for position convergence with the proposed controller in~\eqref{eq:fctrl} and~\eqref{eq:pctrl}. Convergence rates are equal across all axis as expected. It only depends on the gains (equal for all axis) and the mass of the vehicle.

\begin{figure}
  \centering
  \includegraphics[width=0.8\linewidth]{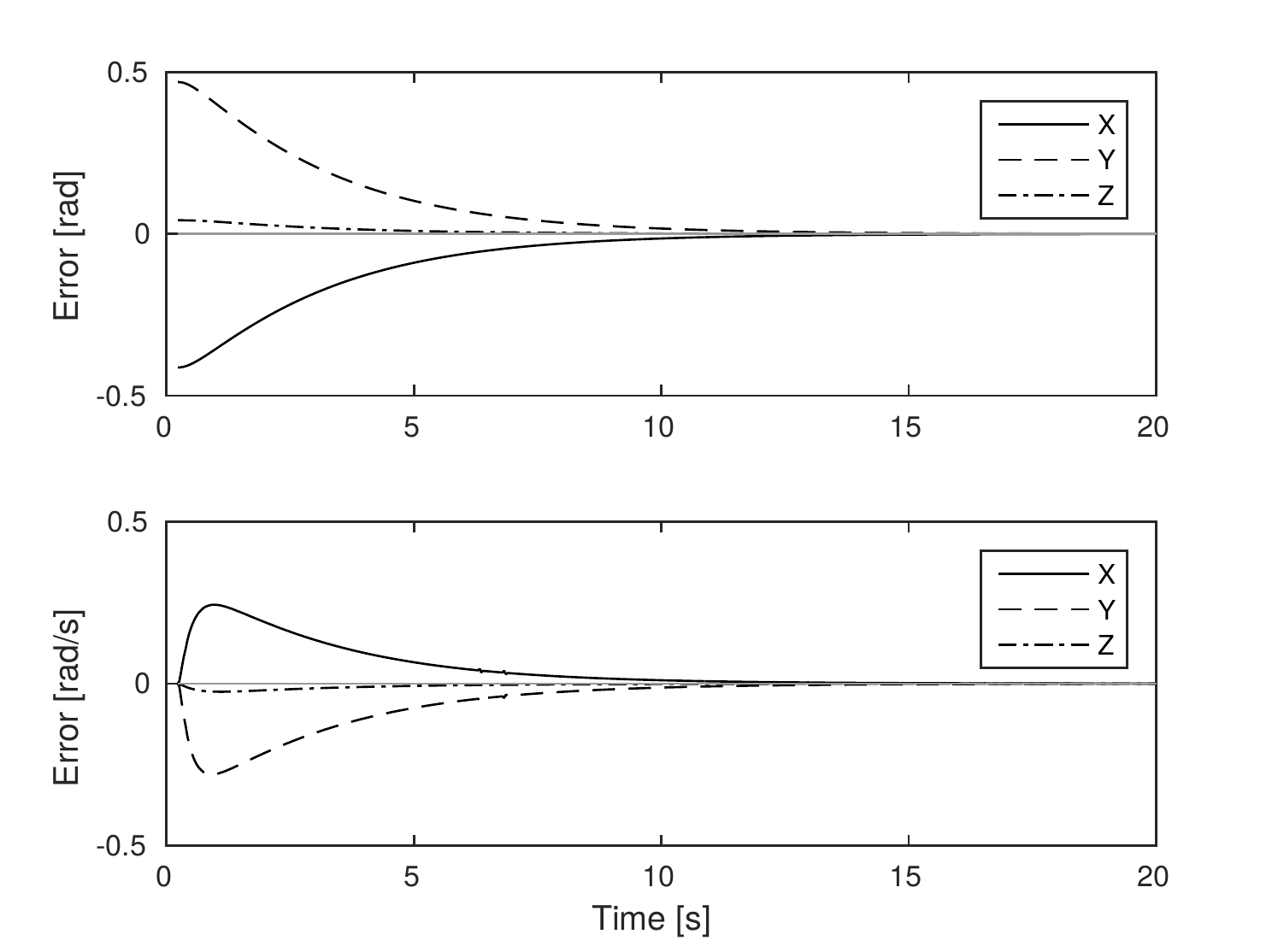}
  \caption{Attitude mode error convergence, from top to bottom: $e_R$ and $e_\omega$.}
  \label{fig:attitude_errors}
\end{figure}

On the attitude simulation, we start the vehicle at $(0\degree,0\degree,0\degree)$ and the reference is set as $(70\degree,-50\degree,30\degree)$ (both in $XYZ$ Euler angles). The results are shown in Figure~\ref{fig:attitude_errors}. Attitude convergence is approximately the same on all axis. As it depends only on the gains (equal for all axis) and the inertia of the vehicle (almost diagonal due to its symmetry), this result was also expectable.

\subsection{Waypoint navigation and payload transportation}

\begin{figure}
  \centering
  \includegraphics[width=0.99\linewidth]{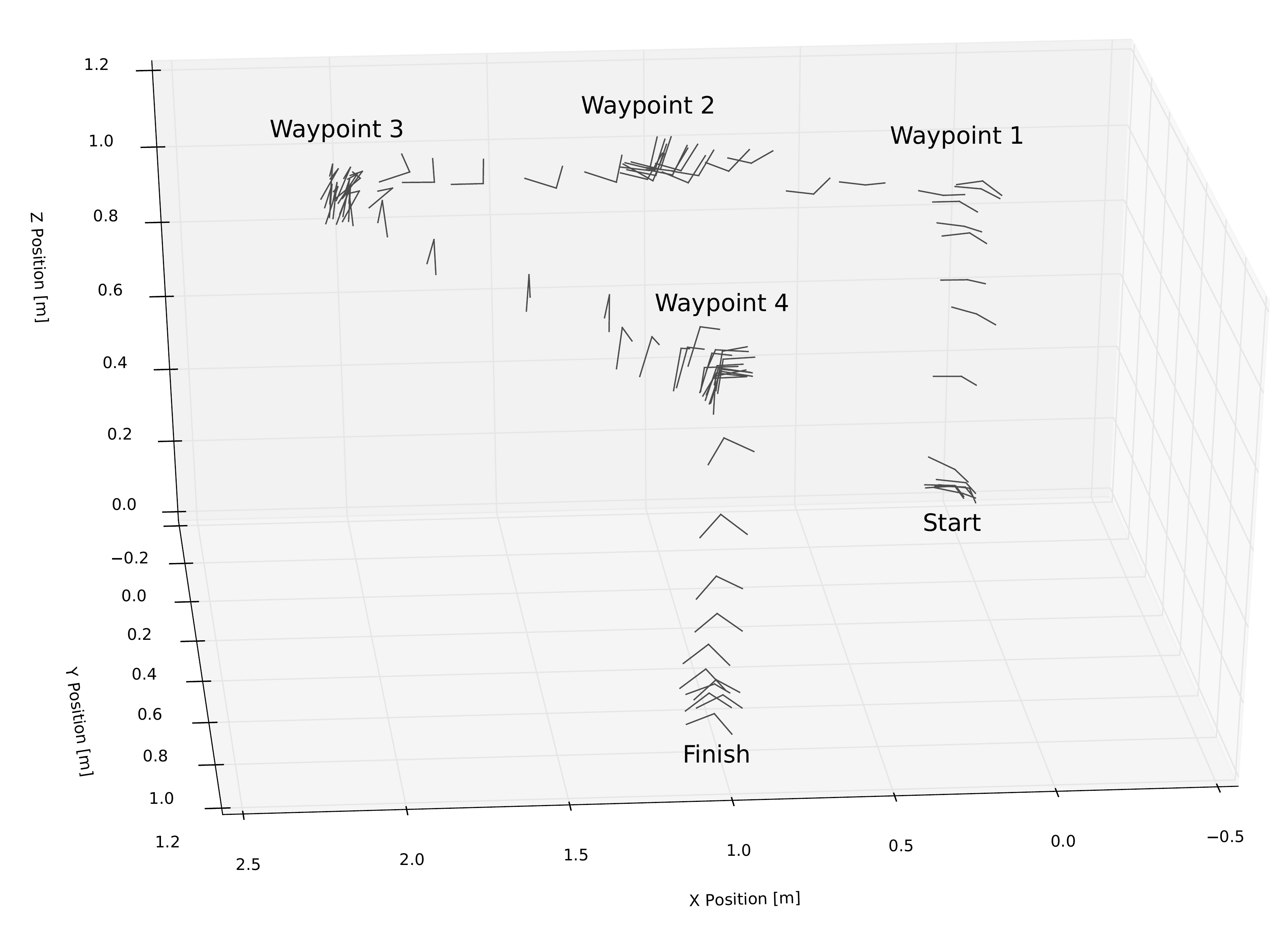}
  \caption{Trajectory followed by the robot without load. Plotted are X and Y axis of the inertia frame~$\mathcal{I}$.}
  \label{fig:trajectory}
\end{figure}

\begin{figure}
  \centering
  \includegraphics[width=0.8\linewidth]{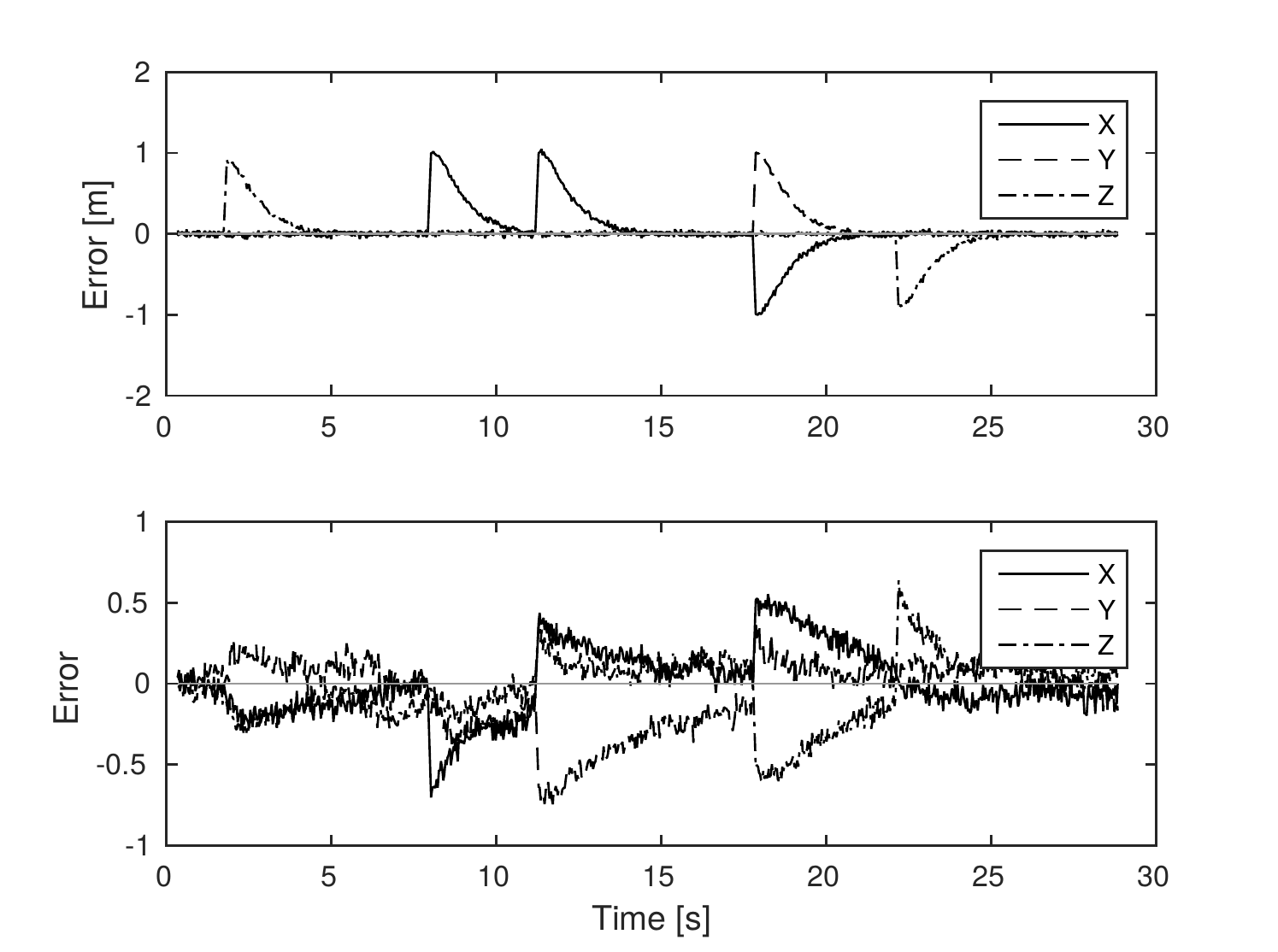}
  \caption{Position (on top) and attitude (on bottom) errors along the no-load trajectory of Figure \ref{fig:trajectory}.}
  \label{fig:nav_errors}
\end{figure}

To validate the joint position and attitude control, we used waypoint navigation: we created a virtual path composed by 6 waypoints, varying on both position and attitude. The trajectory followed by the vehicle is represented on \ref{fig:trajectory}. Plotted are both the X and Y axes of the body frame~$\mathcal{B}$ to represent its attitude. We retrieved the errors across the whole trajectory following, for both position and atitude. These are shown in Figure~\ref{fig:nav_errors}.

\begin{figure}
  \centering
  \includegraphics[width=0.99\linewidth]{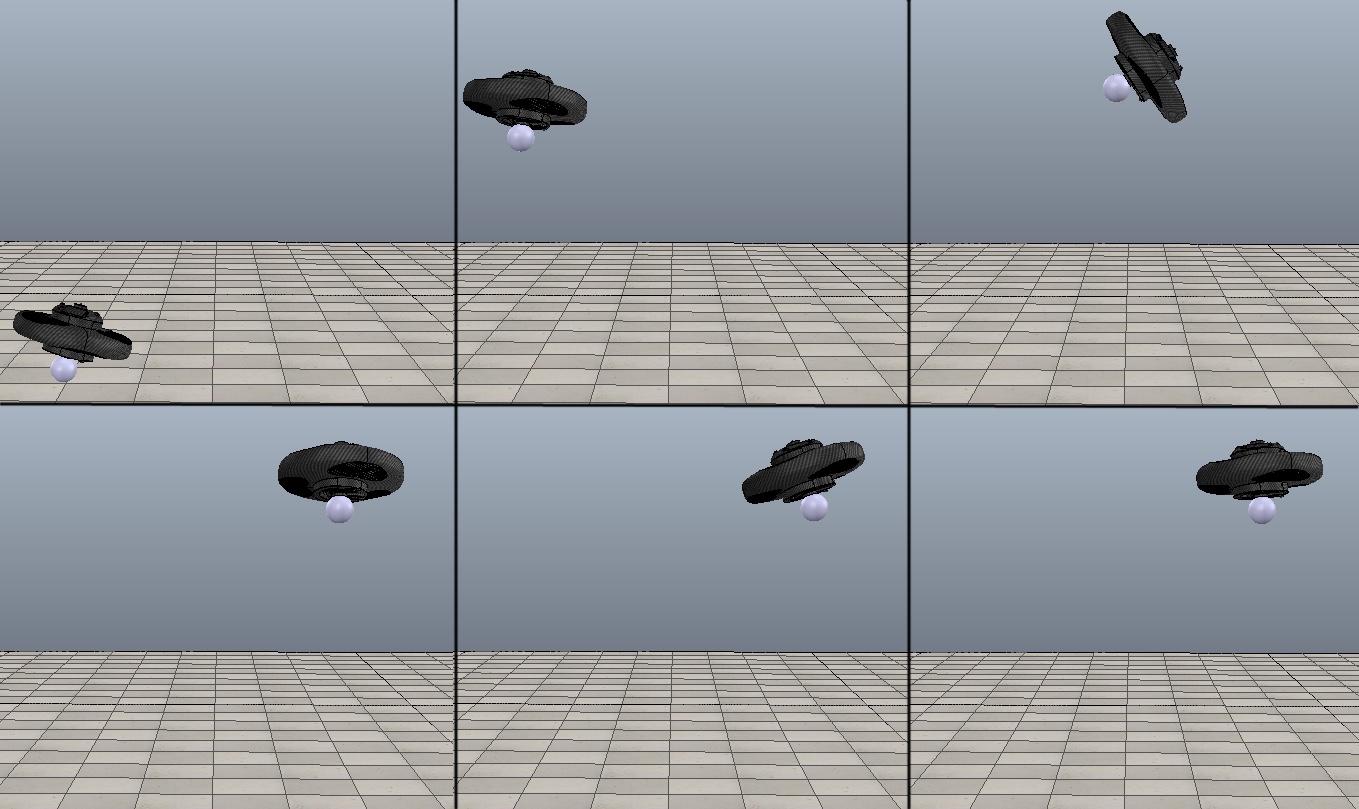}
  \caption{Trajectory followed by the robot with 6kg non-modeled load. Pictured are six positions of the robot along its trajectory.}
  \label{fig:loaded}
\end{figure}

\begin{figure}
  \centering
  \includegraphics[width=0.8\linewidth]{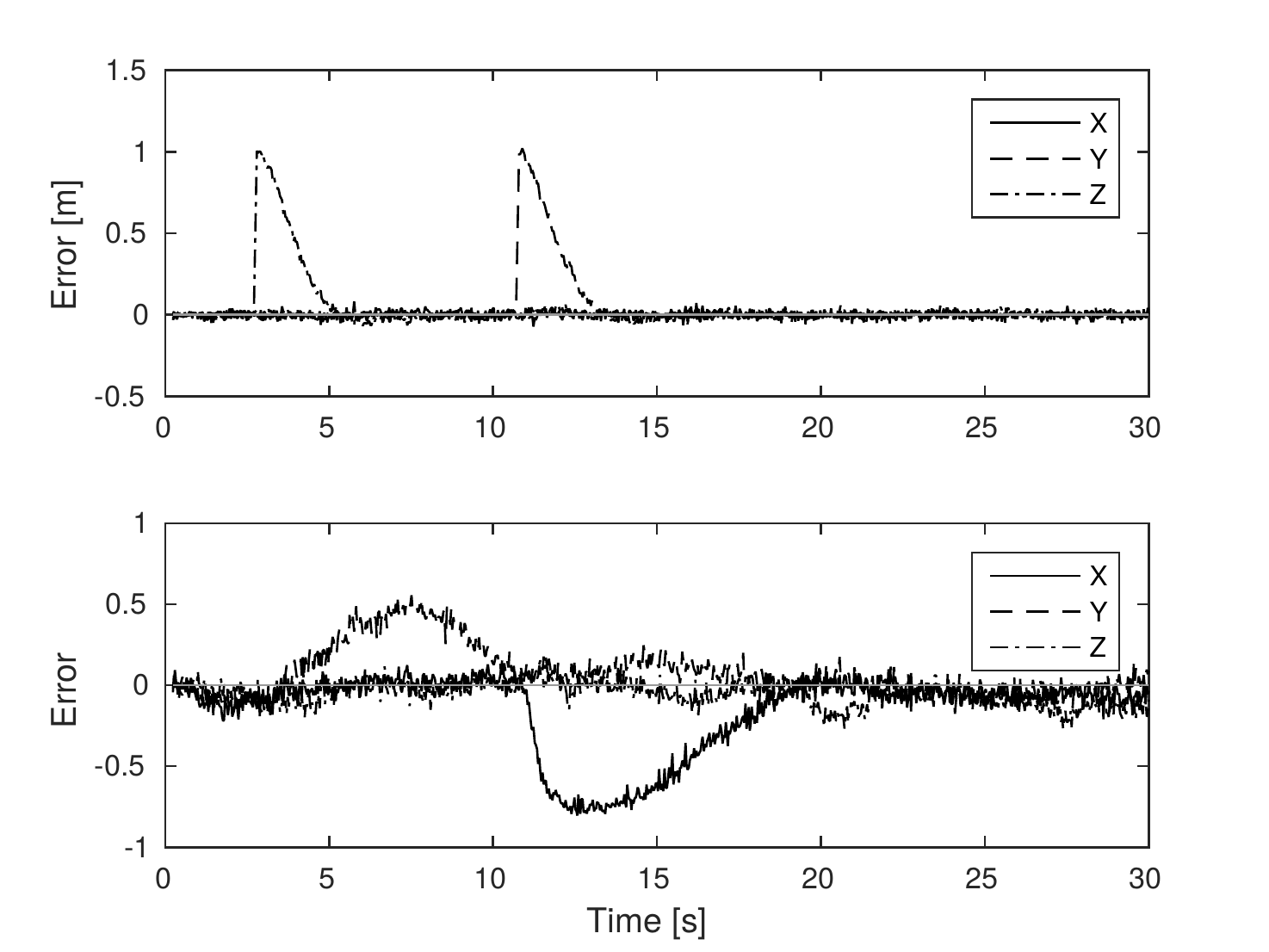}
  \caption{Position (on top) and attitude (on bottom) errors with non-modeled 6kg load along the trajectory of Figure \ref{fig:loaded}.}
  \label{fig:nav_errors_load}
\end{figure}

Then, we added a non-modeled payload to the system: a sphere with 6Kg of mass --- about the same mass of the vehicle. Even with such payload, the controller was able to converge to the required positions, though taking significantly more time. Shown on Figure~\ref{fig:loaded} are screenshots of the vehicle along its 3 waypoint trajectory: 1~meter up along Z and 1~meter right along Y. The position and attitude errors for this trajectory are plotted on Figure \ref{fig:nav_errors_load}

A video showing simulations of Space CoBot in V-REP can be found here: \url{https://youtu.be/M4kdZjxf6-Q} .


\section{CONCLUSIONS AND FUTURE WORK}
\label{sec:concl-future-work}

This paper presented a design of a modular holonomic hexarotor robot for microgravity environments. The propulsion is based on 6 propellers oriented in such a way that both holonomy is attained and the minimum upper bound of the thrust across all directions is maximized. The design is modular, so that maintenance is simplified. We preview as potential applications telepresence and object transportation. The robot can also be used to host microgravity experimental test-beds.

We used a multi-criteria optimization approach to guide the design parameters of the proposed propulsion system. We also include a convergent controller for the vehicle, and a validation of our design in a realistic physics-based simulator. Simulation results show the vehicle navigating between waypoints. In particular, we show that the convergence to the waypoints is robust to localization noise and to unmodeled dynamics, such as attachment to heavy loads.

Future work will address: (1)~vision-based navigation using the onboard camera array, (2)~mobile dexterous manipulation with the onboard robotic arm, encompassing both single robot and cooperative multirobot, and (3)~robustify the controller to better handle unmodeled dynamics, such as heavy loads. We will also aim at constructing a prototype for validating the design both in 2D motion, e.g., frictionless table on Earth, and in microgravity testbeds, e.g., in a parabolic flight aircraft.


\section*{ACKNOWLEDGMENT}

The authors are grateful to André Santos for the CAD modeling of the
vehicle and the renderings presented in this paper. This work was
supported by the FCT project~[UID/EEA/50009/2013].


\bibliography{main}
\bibliographystyle{IEEEtran}

\end{document}